\documentclass[conference]{IEEEtran}
\IEEEoverridecommandlockouts
\usepackage{cite}
\usepackage{amsmath,amssymb,amsfonts}
\usepackage{algorithmic}
\usepackage{graphicx}
\usepackage{array}
\usepackage{textcomp}
\usepackage{xcolor}
\usepackage{tcolorbox}
\def\BibTeX{{\rm B\kern-.05em{\sc i\kern-.025em b}\kern-.08em
    T\kern-.1667em\lower.7ex\hbox{E}\kern-.125emX}}
\begin{document}

\title{PsychFM: Predicting your next gamble\\
}

\author{\IEEEauthorblockN{Prakash Rajan , Krishna P. Miyapuram}
\IEEEauthorblockA{\textit{Cognitive Science \& Computer Science} \\ 
\textit{Indian Institute of Technology, Gandhinagar}\\
Ahmedabad, India\\
\{prakash.r,kprasad\}@iitgn.ac.in}}

\maketitle

\begin{abstract}
There is a sudden surge to model human behavior due to its vast and diverse applications which includes modeling public policies, economic behavior and consumer behavior. Most of the human behavior itself can be modeled into a choice prediction problem. Prospect theory is a theoretical model that tries to explain the anomalies in choice prediction. These theories perform well in terms of explaining the anomalies but they lack precision. Since the behavior is person dependent, there is a need to build a model that predicts choices on a per-person basis. Looking on at the average persons choice may not necessarily  throw light on a particular person’s choice. Modeling the gambling problem on a per person basis will help in recommendation systems and related areas. A novel hybrid model namely psychological factorisation machine ( PsychFM ) has been proposed  that involves concepts from machine learning as well as psychological theories. It outperforms the popular existing models namely random forest and factorisation machines for the benchmark dataset CPC-18. Finally,the efficacy of the proposed hybrid model has been verified by comparing with the existing models.

\end{abstract}


\section{Introduction}

Understanding human behaviour is very much essential in today's world for the top level management  as the existence
of the organisation depends on the employees/individuals/consumers etc.
Developing human behaviour model is challenging as personality, attitudes,values, perception, motives, aspirations and abilities varies from person to person and from time to time. Let us try to understand a typical choice/gamble problem. Consider a user U-1 and some of his previous gamble choices for choice problems are G1 … G25, his gamble choices for next G26 ... G30 needs to be predicted. Note that gamble problems may be different for different users. This task is reminiscent of online recommender systems predicting favourability ratings. This is known as decision making under risk. Let us consider a relatively simple problem first. Given the average choice rate of choice problems, G1 ... G25. The choice rate of a choice problem is the number of times a participant chooses B by the total number of trails. Using average chioce rate the gamble choice rates for next G26 ... G30 gamble problems can be predicted. In this case, the problem is looked from an aggregate behaviour point of view. 

\begin{figure}[h]
    \centering
\begin{tcolorbox}[width=0.3\textwidth]
Gamble A: 3 with certainty \\
Gamble B: 32, .1; 0 otherwise \\
E[A] = 3 E[B] = 3.2  
\end{tcolorbox}
\caption{Example of a choice/gamble problem}
 \label{fig:1}
\end{figure}

An initial approach was to calculate the expected utility function of both the gamble options A and B. There are a lot of anomalies to this theory. As per the example given in figure \ref{fig:1}, 68\% of the participants tend to prefer gamble A over gamble B even though the expected value of gamble B is more than gamble A. This anomaly is called ‘Under-weighting of rare events’ \cite{doi:10.1002/bdm.443}.

In an attempt to address some of these problems, prospect theory was proposed by Kahneman and Tversky \cite{10.2307/1914185}. Kahneman went on to win the Nobel prize in economics for his contribution. The prospect theory addresses the deviation with certainty effect, reflection effect, and also introduces the concept of the value function. With time many more anomalies evolved. Even though they explain the reason for the cause, there is a need for a high precision prediction. Best Estimate and Sampling Tools (BEAST) \cite{Erev2017FromAT} model was developed for this reason. The significance of this model is that it can models 14 such anomalies. 

BEAST model defines the advantage of a gamble over others as the difference between their EV (estimated pessimistically in ambiguous gambles) and the mean value generated by the use of sampling tools that correspond to the four behavioral tendencies. As a result, gamble A will be strictly preferred to gamble B if and only if:

$$ [BEV_A - BEV_B] + [ST_A - ST_B] + e >0 $$

where BEV is the expected value of gambles, ST is the mean value generated by sampling tool, and e is the error term. Psychological features use this model to retrieve different features.

The second line of research is focused on using machine learning models for prediction instead of cognitive models. Cognitive models are theoretical and may work well for small datasets but machine learning models tend to get a slight edge when the dataset gets bigger.

The current state of the art for the aggregate behaviour prediction task is a derived model from BEAST. Most of the high precision models are either derived from BEAST or utilize BEAST at some point of computation.  Psychological forest \cite{Plonsky2016PsychologicalFP} is one such model that uses psychological features derived from BEAST and applies machine learning.

While some of the machine learning models have performed better than theoretical models, the current state of the art uses neural networks with BEAST \cite{Bourgin2019CognitiveMP}. It trains on a synthetic dataset with expected output as BEAST, basically it models neural nets to perform BEAST. Then the neural net is fine-tuned to the competition data and hence performs better than BEAST.

Factorization machines (FM) \cite{Rendle:2010:FM:1933307.1934620} is the current state of the art for the individual behaviour prediction task. FM works well for a sparse input vector. Since this task is similar to online recommendation systems, matrix factorization technique \cite{koren2009matrix}, which won the Netflix challenge, seems to be a good fit.

To meet the aforementioned challenges,the proposed work focused on the hybrid model which evolves from an intersection of two domains - psychological theory and machine learning.

The major contributions of the  paper includes the 1) proposing a new hybrid model which will work with minimal data points and that can outperform all the existing models 2) Carrying out comparative study of the proposed algorithm with all other popularly used models already reported in the literature on a competitive dataset 3)The proposed model and existing models are evaluated on test and validation data to understand the stability of the model.

The rest of the paper is organised as follows. Section 2 presents the CPC18 dataset which is used throughout the paper. It also includes the detailed description of very feature of the dataset. Section 3 discusses the architecture of the machine learning model. Section 4 presents the results for existing and proposed models. Section 5 provides the summary and scope of the proposed model. Section 6 discuss the need and benefits of the hybrid models in behavior modelling frame work.

\renewcommand{\arraystretch}{1.15}
\section{CPC-18: Choice Prediction Competition Dataset}
\subsection{Dataset Description}
A benchmark dataset for evaluating behavior-based decision making is CPC 2018 \cite{plonsky2018and}, which covers a better space of choice problem when compared to its older version CPC 2015\cite{erev2017anomalies}. It covers a wide range of anomalies compared to the other dataset and also is one of the enormous datasets in this domain.

CPC15 contains 150 choice problems, whereas CPC18 contains 210 choice problems. 60 choice problem was added to the CPC18 and the remaining 150 were the same as CPC15. There was a total of 240 participants, out of which 139 are female. Half of the participants came to the Technion and another half at the Hebrew University of Jerusalem. Each participant faced either 30 or 25 choice problems. Each choice problem was conducted for twenty-five repeated trials with feedback for the first five trails. In total, there are 510750 data points. 

\begin{figure}[h]
    \centering
    \includegraphics[width=0.5\textwidth]{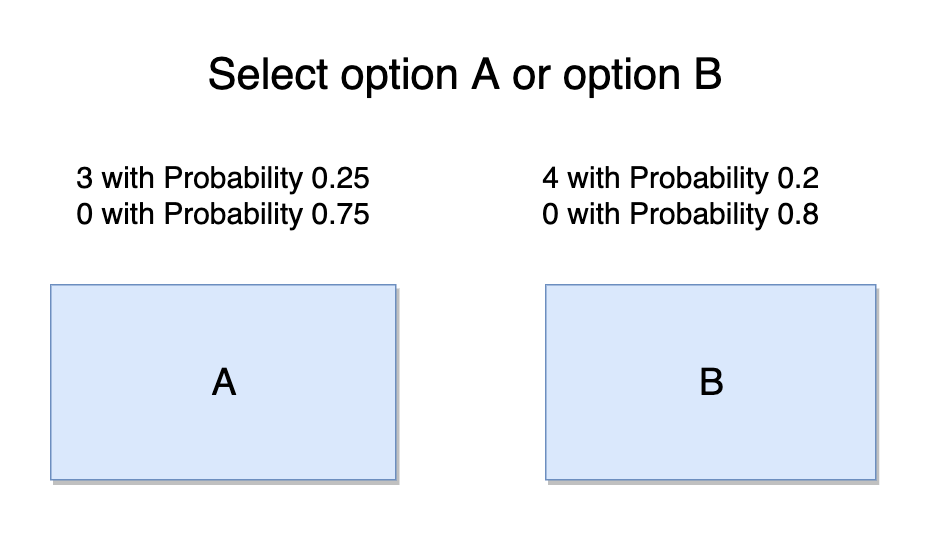}
    \caption{An example of a gamble problem displayed to the user for the CPC dataset.}
    \label{fig:2}
\end{figure}

Participants were paid according to the choice they made. It reduces the noise in data (i.e.) participants will try to choose the gamble choice which he/she thinks will earn him better rewards and rather not choose a random choice. In each problem, participants are faced with two options A and B where gamble A can take up to 2 sets of rewards and gamble B problem was varied by a few parameters such as the distribution. Figure \ref{fig:2} is an example of a gamble presented to a participant. Although in this example the gamble B has only two possible outcomes in general gamble can have more than 2 outcomes.

\begin{figure*}[h]
    \centering
    \includegraphics[width=1\textwidth]{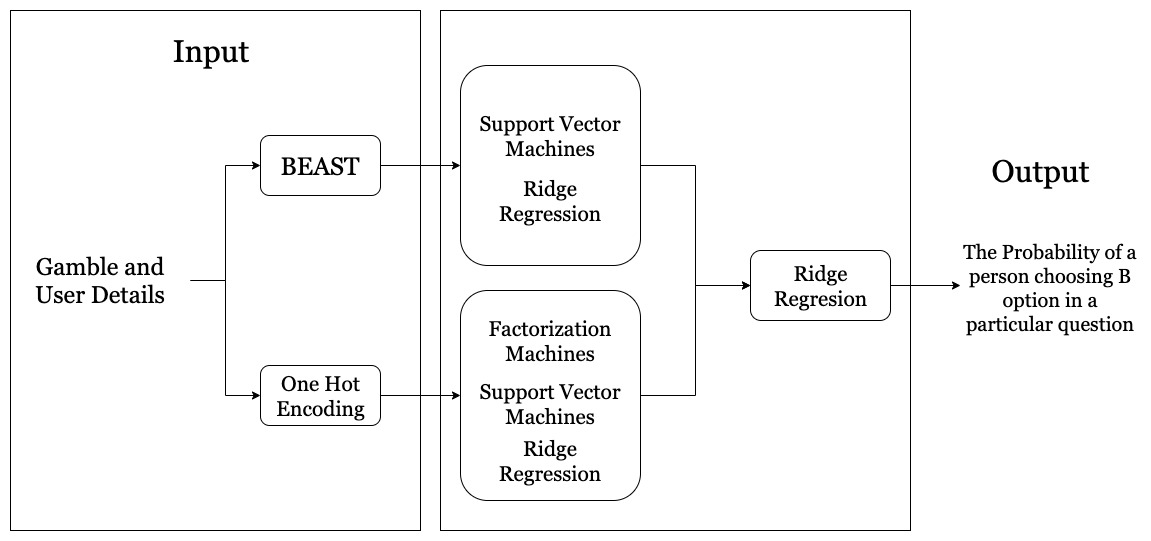}
    \caption{An abstract architecture of the ensemble models implemented in this paper.}
    \label{fig:3}
\end{figure*}

\subsection{Gamble Description}
Each choice problem has a unique set of 11 features. For gamble A, $<$Ha,pHa,La,1-pHa$>$ is defined by three parameters whereas the gamble B is defined as  $<$Hb,pHb,Lb,1-pHb$>$ if the 'LotNum' is 1 else it is $<$Hb,pHb,LotVal,LotShape,LotNum$>$ where 'Lotnum' specifies the number of possible outcomes. 'LotShape' describes the distribution parameter and it can take three values- symmetric, right-skewed, or left-skewed. 'LotVal' is the lottery's expected value. The parameter 'Amb' is set to 1 if the gamble B is ambiguous in the sense that the value of the probability is ambiguous. 

The tenth parameter ' Corr'  captures the correlation between the payoff that the two gambles generate (either positive, negative or none). The 11th parameter ' Feedback'  captures whether Feedback is provided to the decision-maker. As explained above, it is set to 0 in the 1st block of each problem and set to 1 in all other blocks. Each of the 11 parameters that define a problem is provided explicitly to the decision-makers in some way. Feedback is not taken into account for the rest of the paper. Emphasis is given to the average rate of choosing B over all the blocks regardless of feedback.

\subsection{Train-Test Dataset Split}
Out of the available dataset, 5 Random gamble problem per user is chosen to be put into as test set and the others are labeled as train set. For blending, there is a need to create a validation set from the train set so that the ensemble method can learn from the validation set. 10\% of the train set is reserved for the validation purpose.

\section{Model}
The model which surpasses all the existing models are an ensemble of psychological  forest features \cite{Plonsky2016PsychologicalFP} and factorization machines \cite{Rendle:2010:FM:1933307.1934620}. In figure \ref{fig:3}, the architecture of the ensemble model is described. FM performs best when the input vector is sparse. One hot encoded game ID and subject ID are given as input to FM which is very sparse. Even though FM performs well, it does not have the necessary details of the gamble where psychological forest would be of help. In psychological forest, only the features of the gamble are given as input. 

\subsection{One Hot Encoded Vector}
A vector of length 450 is considered in which two of the features are active ones and others are zero. The two active features describe the participant ID and the gamble ID. This is a suitable input for models that perform well on sparse data.

\subsection{Psychological Features}
The features considered for this work includes 11 objective features, 4 naive features and 13 psychological features. Objective features are the features that are already laid out to the participant.
$<$Ha, pHa, … $>$ are the 11 objective features.
Naive features are the ones that lay out some basic comparison between the two gambles and there is no need for psychological theory.  dEV,dSD,dMin,dMax are the difference between the expected value, standard deviation, minimum and maximum possible outcome of gambles respectively.

Table \ref{tab1} illustrates all the 13 Psychological features and their interpretation.

\begin{table}[htbp]
\caption{Psychological features}
\begin{center}
\begin{tabular}{m{2.6cm}|m{5cm}}
\hline
$dEV_{o},dEV_{fb}  $& It describes the difference between the EV of the gambles. It is different from dEV in the sense that it includes the definition of dEV even if the gamble B is Ambiguous.\cite{erev}  \\ 
\hline
$pBetter_{o},pBetter_{fb} $& The probability of gamble B being strictly higher than gamble A. Participants try to minimize the regret. \cite{erev2014maximization}\\ \hline
$dUniEV, pBetter_{u} $& Participants assume the probability of getting any value to be equal. \cite{thorn}\\ \hline
$dSignEV, pBetter_{So},$ $pBetter_{Sfb}$&  Participants give importance to sign. The Ha, La,Hb,Lb values are dropped, and only the sign is taken into account. \cite{payne}\\ \hline
$Signmax$ & An indicator variable. This indicates whether the gamblers have a possibility of positive outcomes. \cite{10.2307/1914185}\\ \hline
$RatioMin $& The ratio between the minimal outcomes. \cite{brandstatter2006priority}\\ \hline
$Dom $& An indicator variable which signals whether a particular gamble dominates.\\ \hline
\end{tabular}
\label{tab1}
\end{center}
\end{table}

\subsection{Factorization Machines}
Factorization machines (FM) are supervised learning models, can do both regression and classification, usually trained by stochastic gradient descent (SGD), alternative least square (ALS), or Markov chain Monte Carlo (MCMC). FM’s are extensions of linear models which model the interactions of variables by mapping the interactions to a low dimensional space. They accomplish this by measuring interactions between variables within large data sets.
 As a result, the number of parameters extends linearly through the dimensions.

$$ \hat{y}(x) = w_o+\sum_{i=1}^{n}w_ix_i +\sum_{i=1}^{n}\sum_{j=i+1}^{n} < V_i , V_j >x_ix_j$$
where, 

$w_o$ is the global bias , 

$w_i$ is the weight of i\textsuperscript{th} feature of the input vector. 

$V_i$ is a vector of dimension k, which represents the i\textsuperscript{th} feature. 

$< V_i , V_j >$ is a vector dot product of the vector representing i\textsuperscript{th} and j\textsuperscript{th} feature.

This model performs extremely well if the data is sparse. To understand why it performs well in a sparse data setup let's take an example.

\textbf{Example 1:} A user U-1 chooses gamble B for a given gamble problem G-1, 21 times out of 25. One hot encoded vector will have $X_1=1 ,X_{350}=1$ others take a 0 value. If this vector is used as input, since there is no interaction between them in the training dataset, most of machine learning models gives $w_{1,350}=0$  whereas the proposed gives $w_{1,350}=< V_1 , V_{350} >$ even though there is no interaction between them in train dataset but there exists an interaction between them and others from which the vectors are updated. $V_i$ is updated every time ith feature is active one. $V_1$ is updated every time the user U-1 chooses a gamble for a gambling problem.

The dimensionality of the hyperplane is defined as k. Linear support vector machine (SVM) is just FM with dimension-1. Hence it fails to gain information about the interactions between features. Fast FM\cite{Bayer:2016:FLF:2946645.3053466} library is used to generate the results presented in this paper. 

\subsection{Ridge and Lasso Regression}
$$ \hat{y}(x) = b+\sum_{i=1}^{n}w_ix_i $$
The above equation models linear regression. To solve for \textbf{w} and \textbf{b} we need to define a cost function first. Let's take the cost function to be Mean Squared Error (MSE).
$$ \text{Cost Function} = \text{MSE} = \sum_{j=1}^m (y_j - \hat{y}_j)^2$$
The optimum solution is obtained by minimizing the cost function. If the error is high for both training and testing data set, then the model is under-fitted and happens when the data set is small. If the error is low for training and high for testing, then the model is over-fitted. Using regularization helps in reducing the over-fitting of the model. Ridge regression is a linear regression with L2 regularization. Lasso regression is a linear regression with L1 regularization.
$$ \text{Ridge Cost Function} = \sum_{j=1}^m (y_j - \hat{y}_j)^2 + \lambda \sum_{j=1}^n w_j^2$$
$$ \text{Lasso Cost Function} = \sum_{j=1}^m (y_j - \hat{y}_j)^2 + \lambda \sum_{j=1}^n |w_j|$$

\subsection{Blending}
Blending is a technique where weighted averaging of predicted output from different model is considered.

$$ \hat{y}(x) = c_1\hat{y_1}(x) + c_2\hat{y_2}(x)$$
Where $\hat{y_1}(x)$ is the prediction from Factorization Machines on one hot encoded input and $\hat{y_2}(x)$ is the prediction from ridge regression on psychological Feature set.

\begin{enumerate}
    \item Divide the dataset into train-Validation sets.
    \item Run the layer-1 models on the train set. Typically these models can be SVM, multilayer perceptron (MLP) and linear regression, etc.
    \item The input of layer 2 is the prediction of layer-1 models on the validation set.
    \item Run the layer-2 models on the validation set.
\end{enumerate}

Ridge regression is used to determine the coefficients $ c_1 $ and $c_2 $. From the coefficients,  the significance of the model can be determined. If one model's coefficient is significantly greater than the other, then blending the models does not improve the accuracy significantly.

\begin{table}[htbp]
\caption{Mean Squared Error of Different Models}
\begin{center}
\begin{tabular}{l c}
& \textbf{MSE*100} \\
\textbf{Naive Models on One Hot Encoded Input (A)}& \\
\hspace{1em}Factorization Machines & \textbf{7.63}\\
\hspace{1em}Ridge Regression & 8.36\\
\hspace{1em}MLP (200,50,10) & 10.4 \\
\hspace{1em}SVM & 12.32\\
\hspace{1em}Lasso Regression & 13.88\\
\hline 
\textbf{Naive Models on Psychological Feature Input (B)} & \\
\hspace{1em}Ridge Regression & \textbf{7.80}\\
\hspace{1em}Lasso Regression & 7.99\\
\hspace{1em}SVM & 14.90\\
\hspace{1em}Random Forest & 14.97\\
\hspace{1em}MLP (10,2) & 
17.2 \\
\hline
\textbf{Ensemble Models} & \\
\hspace{1em}FM (A) + Ridge (B) & \textbf{6.8} \\
\hspace{1em}Ridge (A) + Ridge (B) & 7.2 \\
\hspace{1em}MLP (A) + Ridge (B)& 7.8 \\
\hspace{1em}FM (A) + Lasso (B)& 7.9 \\

\hline
\end{tabular}
\label{tab2}
\end{center}
\end{table}
\begin{table}[htbp]
\caption{Validation and Test Errors}
\begin{center}
\begin{tabular}{m{3cm}|m{2cm}|m{2cm}}
Model & Test MSE & Validation MSE \\
\hline
Lasso (B) & 7.99 & 19.24\\
\hline
SVM (B) & 14.90 & 27.48\\
\hline
Ridge (B) & 7.8 & 7.63 \\
\hline
FM (A) & 7.63 & 7.42 \\
\end{tabular}
\label{tab3}
\end{center}
\end{table}

\section{Results}
\begin{figure}[h]
    \centering
    \includegraphics[width=0.5\textwidth]{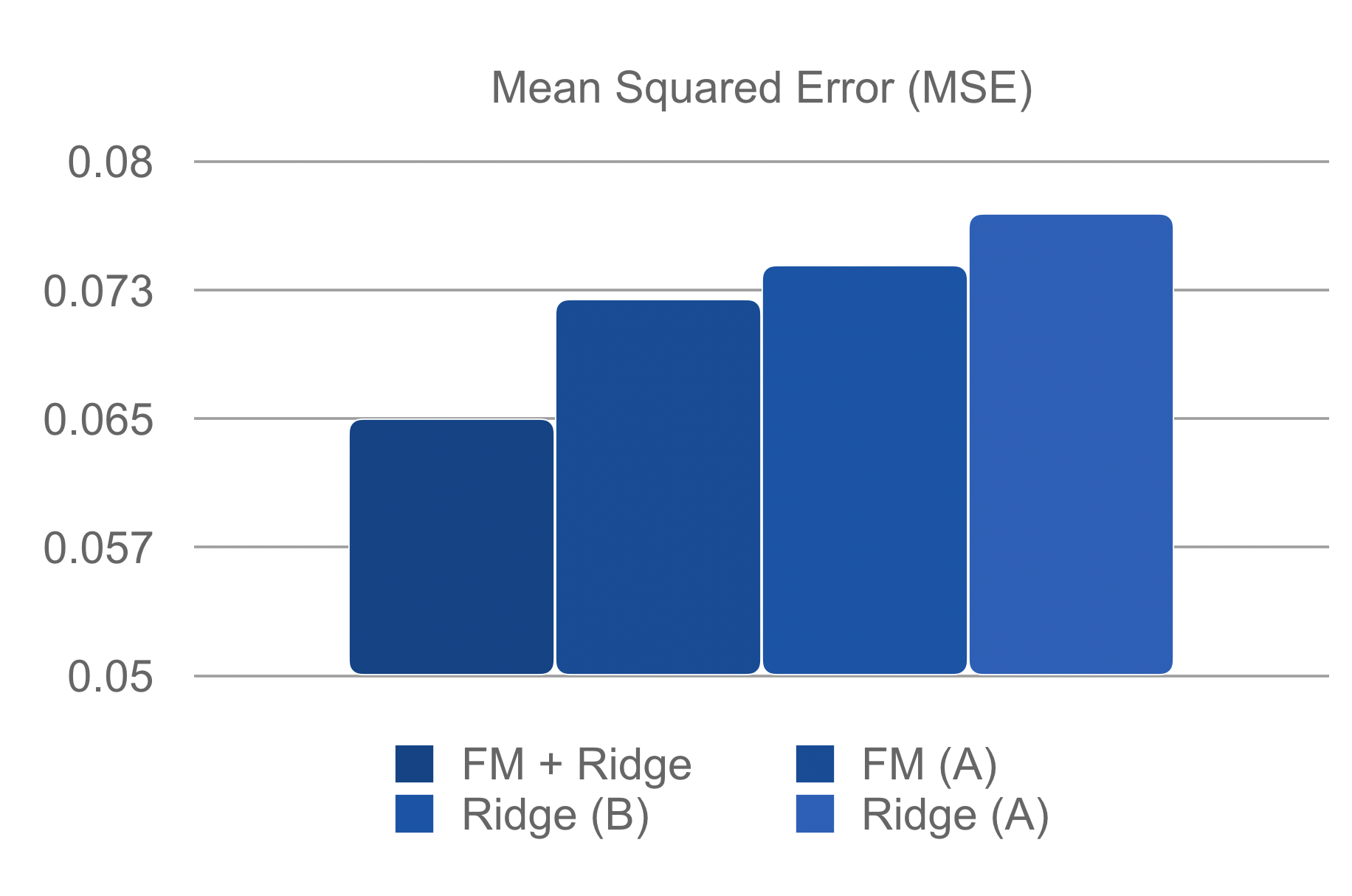}
    \caption{MSE of Top 4 performing models.}
    \label{fig:4}
\end{figure}

A vast number of machine learning models are applied to this problem with two types of input - one hot encoded input and psychological features input. The MSE (Mean Squared Error) of different models with respective inputs is listed in Table \ref{tab2}. In figure \ref{fig:4}, the results of the top performing models are shown graphically. For input type A - one-hot encoded vector, FM models outperform other machine learning models by a fine margin. FM models perform better well due to the level of sparsity in the data. FM models by considering all the interaction between the feature whereas SVM does not.

FM achieves an MSE of 0.0736. On an average for one prediction there is an error of 0.27. That is, let us take the probability of a given person choosing a gamble B for a particular gamble problem is p. FM on an average, predicts the probability as p $\pm$ 0.27. There is a 27\% error in the predicted probability. Surprisingly MLP  performs better than SVM and lasso regression. Ridge performs way better than lasso which signifies the importance of regularization in machine learning models.

For input type B - psychological features, ridge regression performs the best. Close comes lasso regression. The main take away is that linear regression performs well in this setting. The MLP(10,2) means the neural net has 2 hidden layers with 10 and 2 neurons. MLP(200,50,10) means the neural net has 3 hidden layers with 200, 50 and 10 neurons respectively. 

Ensemble method combines one method from the input type A model and one from the input type B model. The least MSE was achieved when FM and ridge are blended. Most combinations of models from input type A and input type B performed better than combining models from the same input type. Due to the fact that one feature is independent of the other,one contains only details of user id and gamble id whereas other has only the details of a gamble. There is no intersection between the two input feature set. Hence they outperform most of the other models.

\begin{figure}[h]
    \centering
    \includegraphics[width=0.35\textwidth]{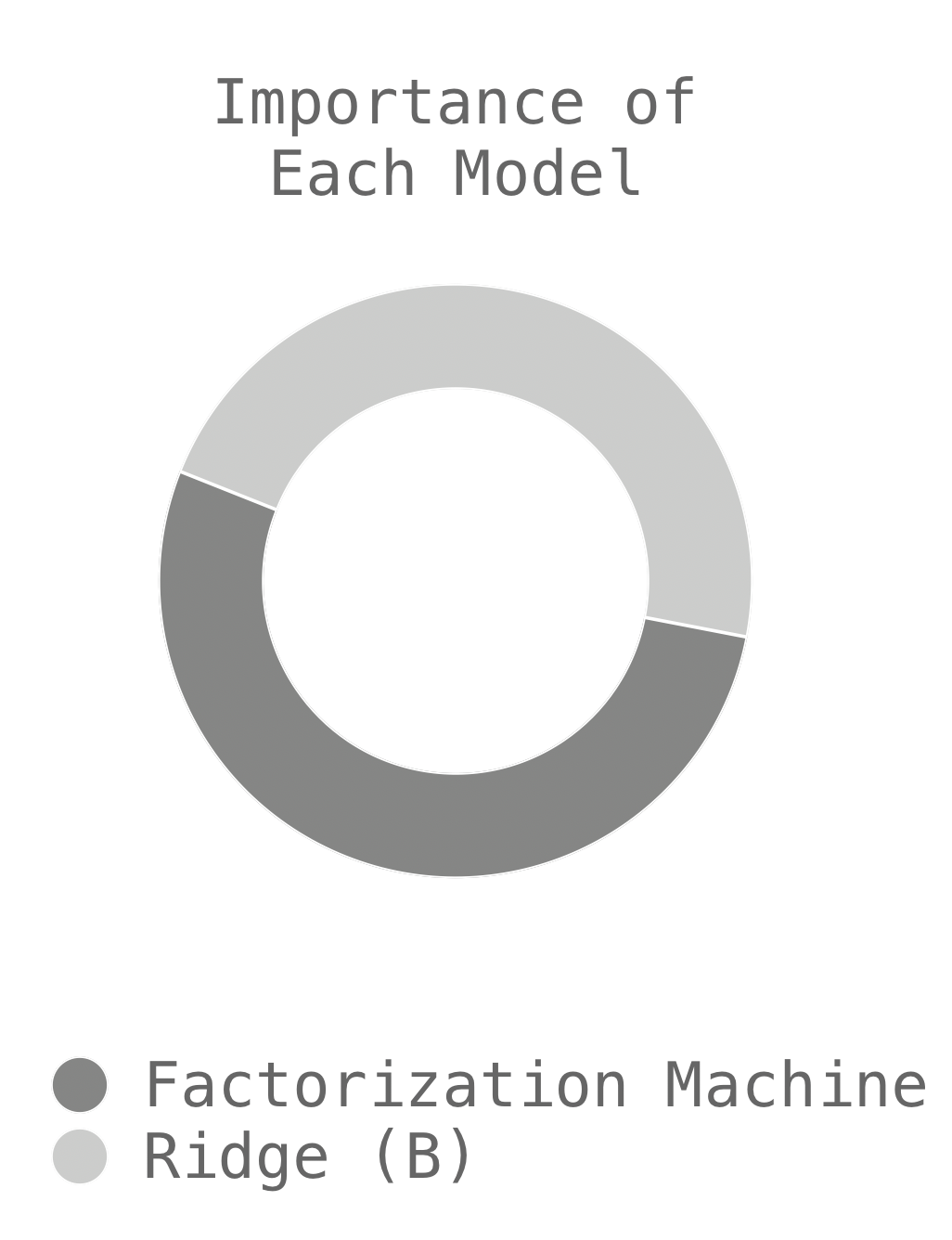}
    \caption{Importance of each model in FM + Ridge ensemble model.}
    \label{fig:5}
\end{figure}

The stability of these models can be inferred from their errors in the validation set and test set. If both of the errors are close to each other then the model is stable. That is, if the error for one set of inputs varies with another set, that is means the model is either under-fitted or over-fitted. In table \ref{tab3}, the error in validation and test set are given for various models. Lasso and SVM have varying errors which specifies the models are not stable, and hence there are not the best model for this setup. In contrast, ridge and FM models have a similar error which inference that the model is stable in this setup. 

The figure \ref{fig:5} shows the importance of each model in ensemble of FM + Ridge. The importance is calculated by the coefficient of blending, that is ,
$$ \hat{y}(x) = 0.529*\hat{y_1}(x) + 0.463*\hat{y_2}(x)$$
$$ c_1 = 0.529$$
$$ c_2 = 0.463$$

In the case of the best performing model, the FM contributes 53\%, and ridge contributes 47\%. That shows both the model are essential to achieve a better performing model than naive models. Whenever more models are added for blending, the error does not decrease significantly or sometimes even increased. The contribution made by the newly added model was also significantly low. Adding more models will also increase the complexity of the model. Thus in this paper, the number of models to be blended is limited to two.

\section{Summary}
\begin{itemize}
    \item FM models are a good fit for high dimensionally sparse data. Typically used for one hot encoded inputs.
    \item Blending FM model and Ridge(B) gives a highly precise model. Logically because the ensemble factors in the user's history and the present gamble's details.
    \item There is a high variance in error with different regularization techniques. Choosing a model with suitable regularization is essential for a high precision system.
\end{itemize}
\section{Discussion}
An amalgamation of cognitive methods and data science model outperform most of the best practices in the CPC dataset. Cognitive model's output remains the same irrespective of external factors. Let’s say these CPC experiments were conducted in a well-developed country versus a developing country that is under economic decline. Since participants are paid based on their performance, participants in a developed country may take more risks when compared to ones from developing countries. Cognitive models may not perform the best since they can’t factor in these external factors whereas in data science models, they look at the data and provide the best fit possible. 

Cognitive models do tend to perform well if the dataset is small since there is not much for the data science model to learn from whereas when the dataset is large, the data science model does well. Hybrid models take the best out of the two worlds. 


\bibliographystyle{IEEEtran}
\bibliography{IEEEabrv,IEEEexample}

\end{document}